\documentclass{sig-alternate-05-2015}
\usepackage{caption}
\usepackage{subfigure}
\usepackage{graphicx}
\usepackage{rotating}
\usepackage{multirow}
\usepackage[usestackEOL]{stackengine}
\setstackgap{L}{\normalbaselineskip}
\newcommand{\norm}[1]{\left\lVert#1\right\rVert}
\begin{document}


\CopyrightYear{2018}
\setcopyright{rightsretained}
\conferenceinfo{SAC 2018}{April 9--13, 2018, Pau,	France}
\isbn{978-1-4503-5191-1/18/04}
\doi{https://doi.org/10.1145/3167132.3167379}

\title{Deep metric learning for multi-labelled radiographs}
%
%
%
%
%

\numberofauthors{4} 
%
\author{
%
%
 \alignauthor 
 Mauro Annarumma\\
        \affaddr{Department of Biomedical Engineering}\\
        \affaddr{King's College London}\\
        \email{mauro.annarumma@kcl.ac.uk}
 \and
 \alignauthor 
 Giovanni Montana\\
        \affaddr{Department of Biomedical Engineering}\\
        \affaddr{King's College London}\\
        \email{giovanni.montana@kcl.ac.uk}
 }

\maketitle
\begin{abstract}

Many radiological studies can reveal the presence of several co-existing abnormalities, each one represented by a distinct visual pattern. In this article we address the problem of learning a distance metric for plain radiographs that captures a notion of ``radiological similarity": two chest radiographs are considered to be similar if they share similar abnormalities. Deep convolutional neural networks (DCNs) are used to learn a low-dimensional embedding for the radiographs that is equipped with the desired metric. Two loss functions are proposed to deal with multi-labelled images and potentially noisy labels. We report on a large-scale study involving over 745,000 chest radiographs whose labels were automatically extracted from free-text radiological reports through a natural language processing system. Using 4,500 validated exams, we demonstrate that the methodology performs satisfactorily on clustering and image retrieval tasks. Remarkably, the learned metric separates normal exams from those having radiological abnormalities. 
\end{abstract}

%
%

\ccsdesc[500]{Computing methodologies~Dimensionality reduction and manifold learning}
\ccsdesc[500]{Computing methodologies~Neural networks}
\ccsdesc[300]{Computing methodologies~Visual content-based indexing and retrieval}
\ccsdesc[100]{Applied computing~Imaging}

%
%

%
%
\printccsdesc


\keywords{deep metric learning, convolutional networks, x-rays}

	\section{Introduction} \label{intro}
	
	Chest radiographs are performed to diagnose and monitor a wide range of conditions affecting lungs, heart, bones, and soft tissues. Despite being commonly performed, their reading is challenging and interpretation discrepancies can occur. There is a need to develop machine learning algorithms that can assist the reporting radiologist. In this work we address the problem of learning a distance metric for chest radiographs using a very large repository of historical exams that have already been reported. An ideal metric should be able to cluster together radiographs presenting similar radiological abnormalities and place them far away from exams with normal radiological appearance. Learning a suitable metric would enable a variety of applications, from automated retrieval of radiologically similar exams, for teaching and training, to their automated prioritization based on visual patterns. 
	
	The problem we discuss here is challenging for several reasons. First, the number of potential abnormalities that can be observed in a chest radiograph can be quite large. Visual patterns detected in radiographs are important cues used by the clinicians when making a diagnosis. Often, during the reporting time, the clinician will describe the visual pattern using descriptors (e.g. ``enlarged heart") or stating the exact medical pathology associated with the visual pattern (e.g. ``consolidation in the right lower lobe"). A metric learning algorithm should be able to deal with any such labels and their potential overlaps. Second, the labels may not always be accurate or comprehensive due to the fact that not all the abnormalities are always reported in an image, e.g. due to omissions or when deemed unimportant by the radiologist. When these labels are automatically obtained from free-text reports, as we do in this work, mislabelling errors may also occur. Third, certain abnormalities are less frequently observed than others, and may not even exist in the training dataset. 
	
	To support this study, we have prepared a large repository consisting of over $745,000$ chest radiograph examinations extracted from the PACS (Picture Archiving and Communication System) of a large teaching hospital in London. To our knowledge, this is the largest chest radiograph repository to ever be deployed in a machine learning study. Due to the large sample size, manual annotation of all the exams is unfeasible. All the historical free-text reports have been parsed using a Natural Language Processing (NLP) system, which has identified and classified any mention of radiological abnormalities. As a result of this process, each film has been automatically assigned to one or multiple labels. Our contributions are the following. First, we discuss the problem of deep metric learning with multi-labelled images and propose two versions of a loss function specifically designed to deal with overlapping and potentially noisy labels. At the core of the architecture, a DCN is used to learn compact image representations capturing the visual patterns described by the labels. Second, we report on a large-scale evaluation of the proposed methodology using a manually curated subset of over $4,500$ exams. Each historical radiological report was reviewed by two independent clinicians who extracted all the labels associated to the films. We report on comparative results for two tasks, clustering and image retrieval, and provide evidence that the learned metric can be used to cluster radiographs with a normal appearance as well as clusters of abnormal exams with co-occurring abnormalities.  
	
	
	\begin{figure}[t]
		\centering
		\begin{minipage}[b]{.49\textwidth}
			\centering
			\begin{minipage}[b]{.5\textwidth}
				\begin{minipage}[b]{1\textwidth}
					\includegraphics[width=1\linewidth]{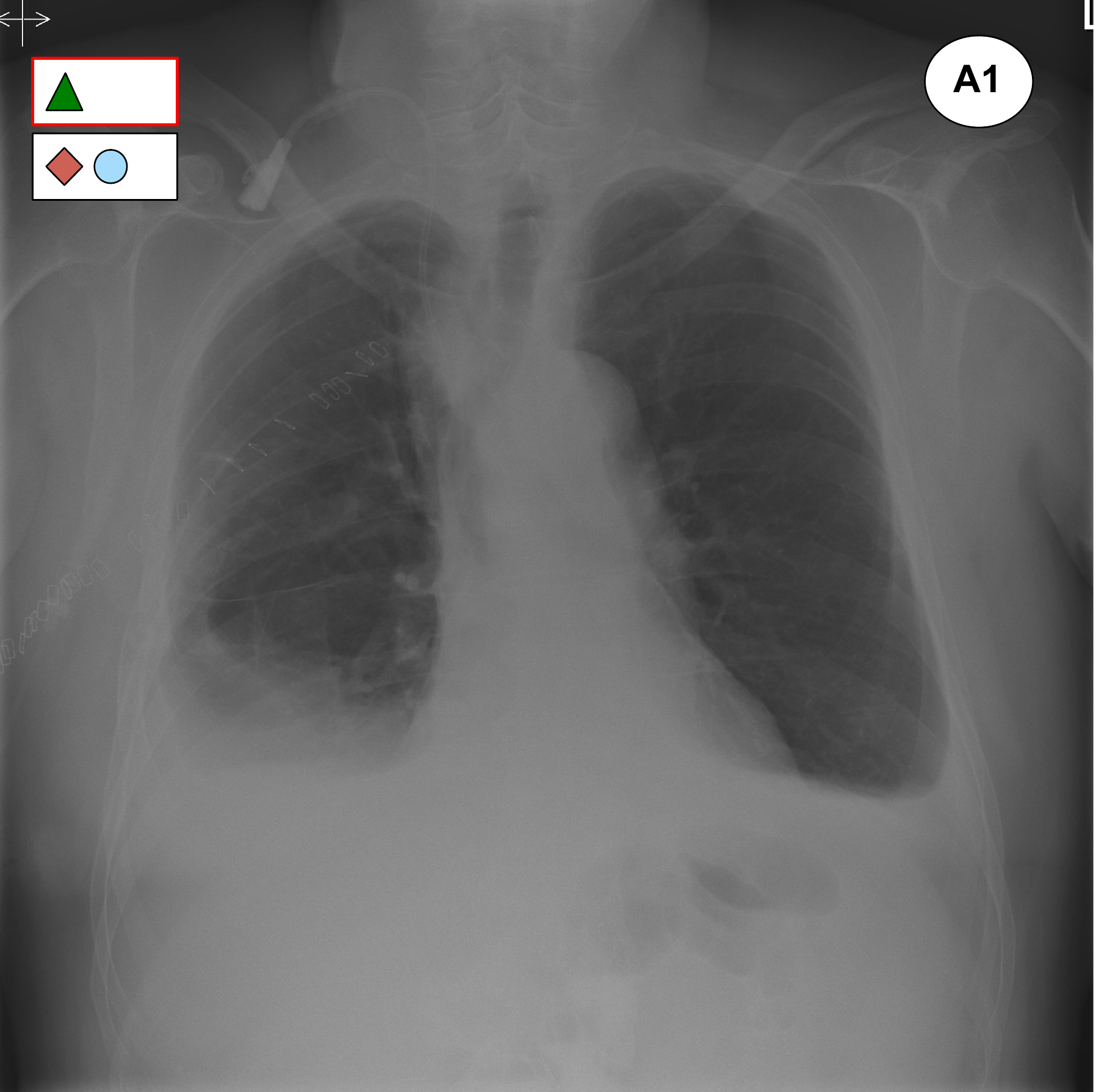}
				\end{minipage}
				\begin{minipage}[b]{1\textwidth}
					\includegraphics[width=1\linewidth]{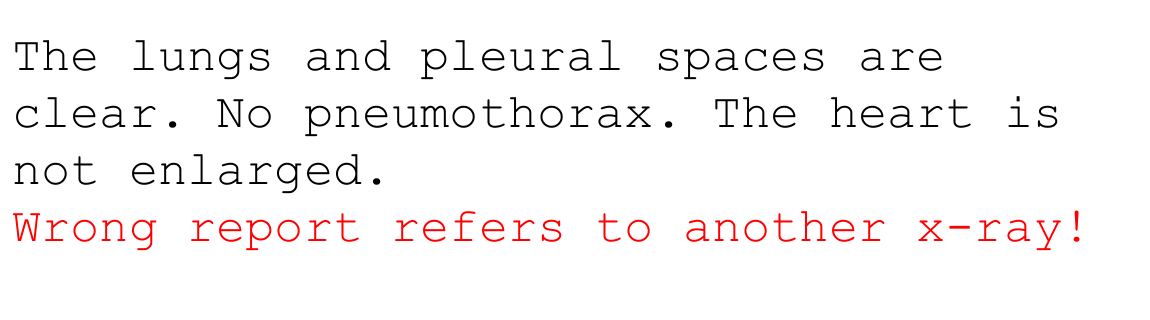}
				\end{minipage}
			\end{minipage}%
			\begin{minipage}[b]{.5\textwidth}
				\begin{minipage}[b]{1\textwidth}
					\includegraphics[width=1\linewidth]{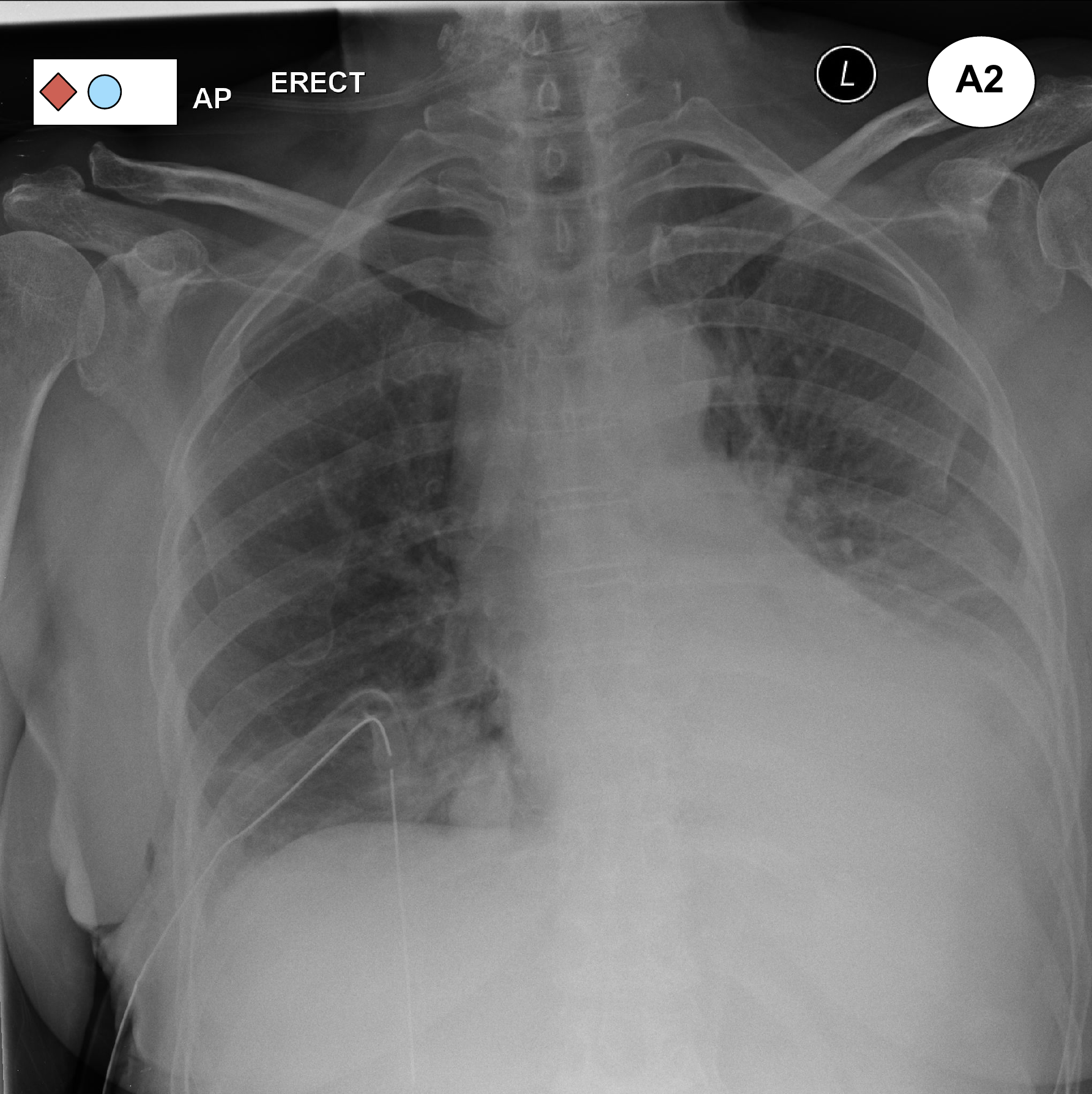}
				\end{minipage}
				\begin{minipage}[b]{1\textwidth}
					\includegraphics[width=1\linewidth]{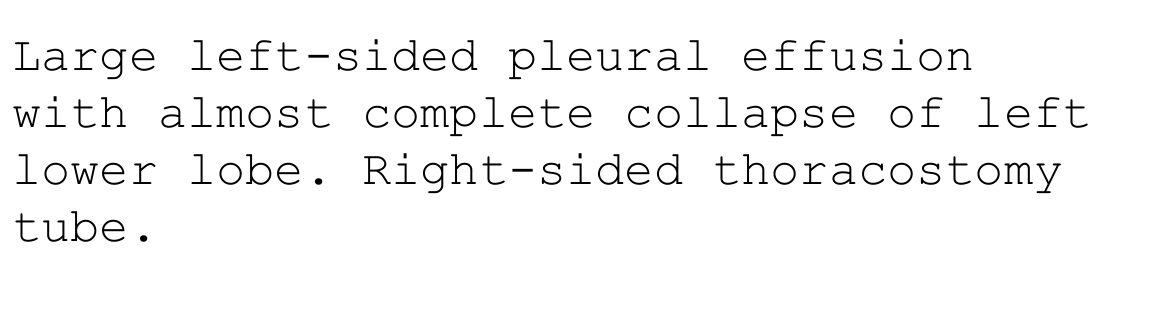}
				\end{minipage}
			\end{minipage}		
			
		\end{minipage}%
		\hfill
		\begin{minipage}[b]{.49\textwidth}
			\centering
			\begin{minipage}[b]{.5\textwidth}
				\begin{minipage}[b]{1\textwidth}
					\includegraphics[width=1\linewidth]{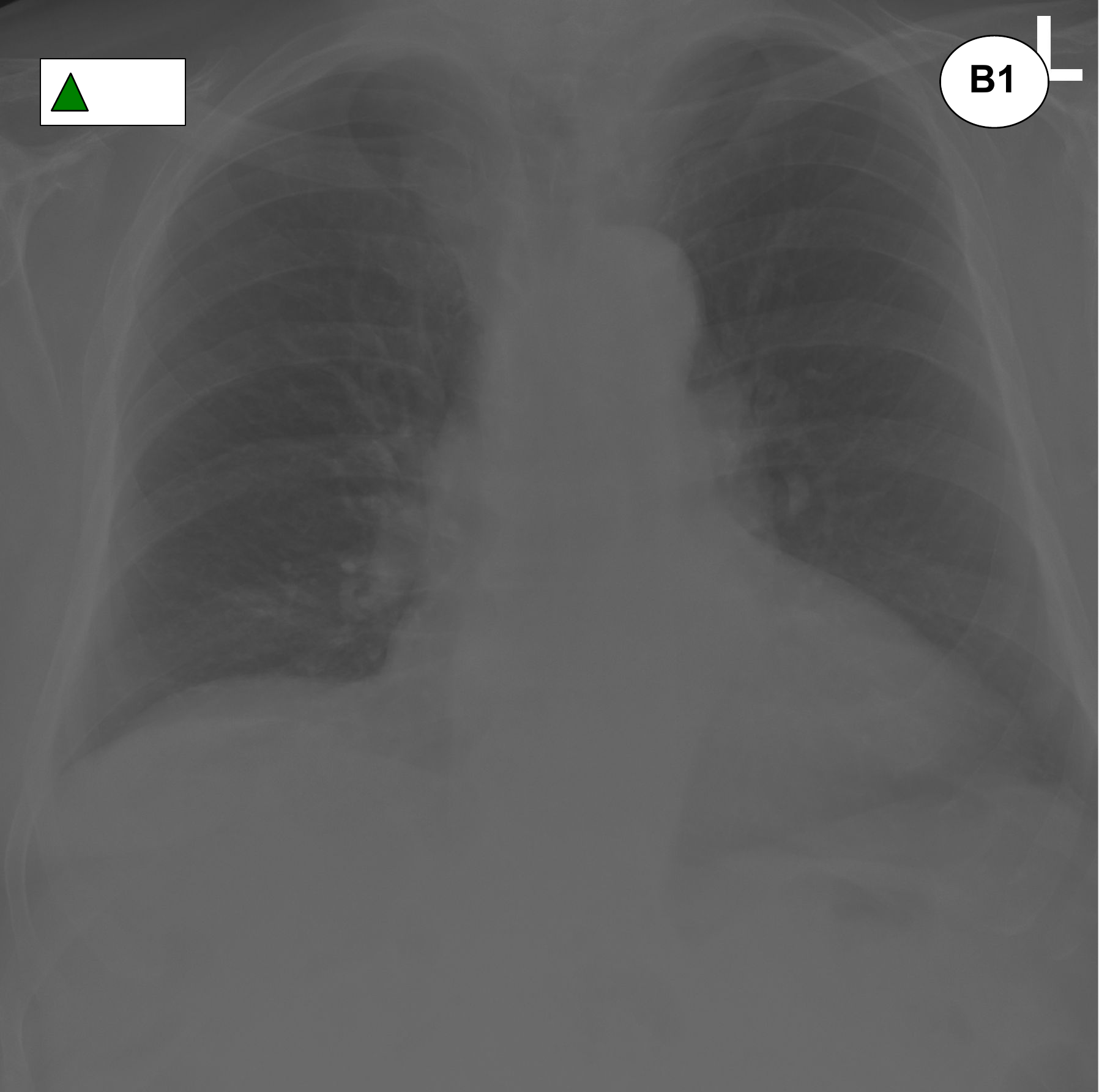}
				\end{minipage}
				\begin{minipage}[b]{1\textwidth}
					\includegraphics[width=1\linewidth]{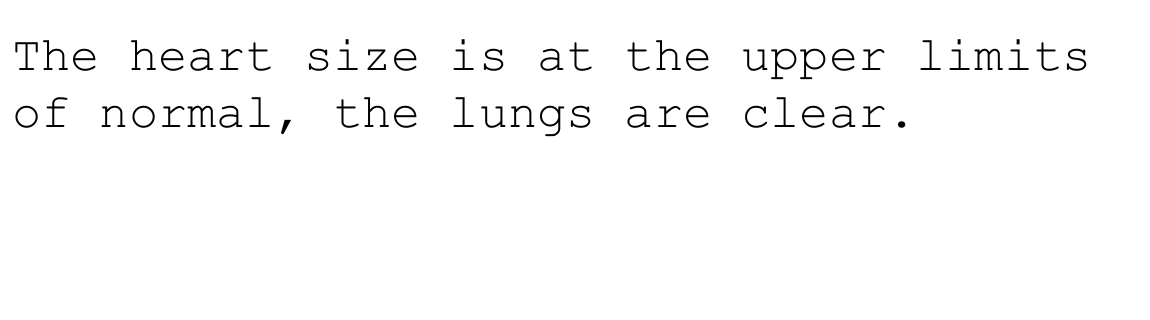}
				\end{minipage}		
			\end{minipage}%
			\begin{minipage}[b]{.5\textwidth}			
				\begin{minipage}[b]{1\textwidth}
					\includegraphics[width=1\linewidth]{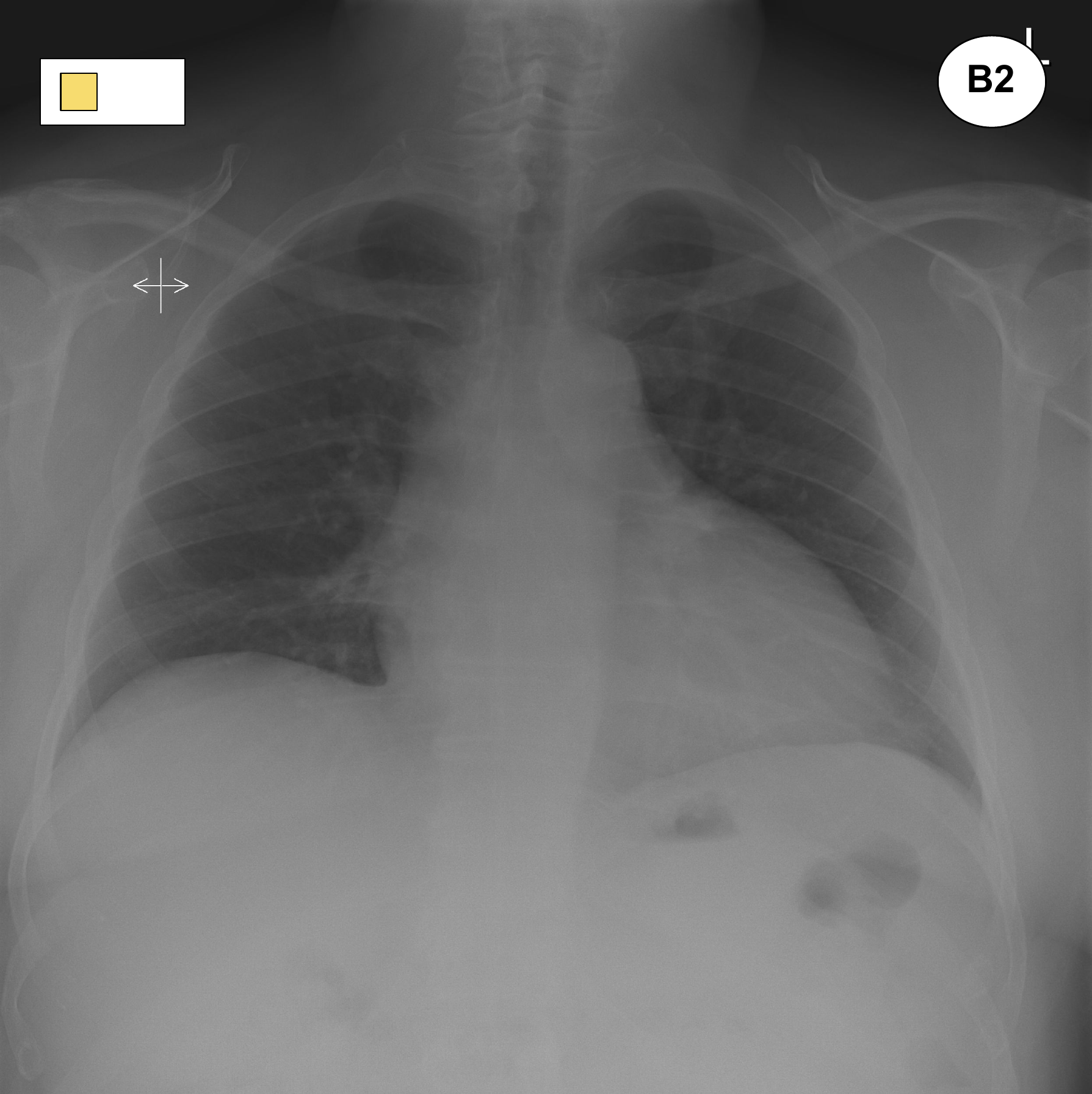}
				\end{minipage}			
				\begin{minipage}[b]{1\textwidth}
					\includegraphics[width=1\linewidth]{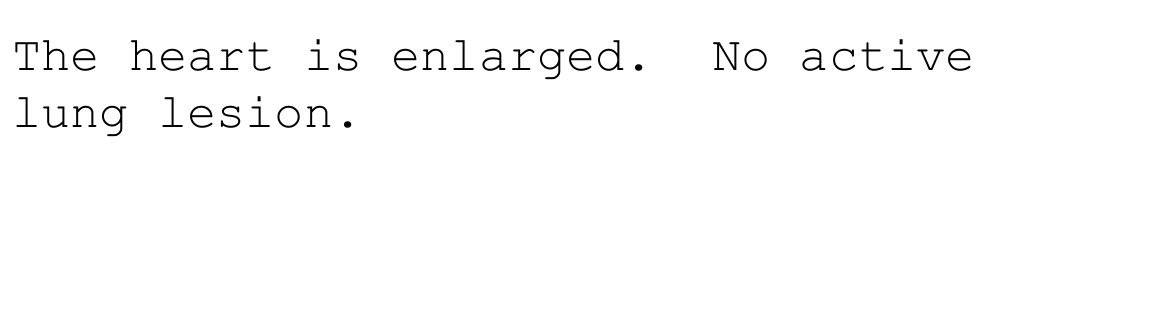}
				\end{minipage}
			\end{minipage}		
			
		\end{minipage}	
		
		\caption{Examples of pairs of images that are placed close to each other in the learned embedding space shown in Fig. \ref{fig:TSNE}. A1 was incorrectly reported, but a second reading shows the presence of pleural effusion and a medical device, which justifies its proximity to A2. B1 was labelled as ``normal'', but a second reading reveals some degree of cardiomegaly and, as such, the scan is placed close to B2. An extract from the original reports can be found under each image. Fig. \ref{fig:TSNE} contains the legend for the labels.}
		\label{fig:x-rays}
	\end{figure}

	\section{Related work}
	\label{sec:related_works}
	
	\subsection{Deep metric learning}
	
	The first attempt of using neural networks to learn an embedding space was the \textit{Siamese Network} \cite{NIPS1993_769}\cite{chopra2005learning}, which used a contrastive loss to train the network to distinguish between pairs of examples. Schroff et al. \cite{schroff2015facenet} combined a Siamese architecture with a triplet loss\cite{NIPS2005_2795} and applied the resulting model to the face verification problem obtaining a nearly human performance. Other approaches have been proposed more recently in order to better exploit the information in each mini-batch; e.g. Song et al. \cite{song2015deep} proposed a loss with a lifted structure, while Sohn et al. \cite{sohn2016improved} proposed a tuplet loss. They both use all the possible example pairs within each mini-batch. All these methods use a query or anchor image $x^a$, which is compared with \textit{positive} elements (images sharing the same label) and \textit{negative} elements (images with a different label). Several of these methods also implement a {\it hard data mining} approach whereby samples within a given pair or triplet are selected in such a way to represent the hardest positive or negative example with respect to the given anchor. This strategy improves both the convergence speed and the final discriminative performance. In FaceNet \cite{schroff2015facenet}, pairs of anchor and positive samples are randomly selected while negative samples are selected from a subset of the training set using a semi-hard negative algorithm. 
	Recently, Wu et al. \cite{samplingmatters_WuMSK17} proposed a novel off-line mining strategy that, on the entire training set, selects the optimal positive and negative elements for each anchor. A different learning framework that does not require the training data to be processed in paired format has been recently proposed \cite{Song_CVPR2017}. 

	\begin{figure*}[t]
		\centering
		\begin{minipage}[b]{.22\textwidth}
			\centering			
			\includegraphics[width=1\linewidth]{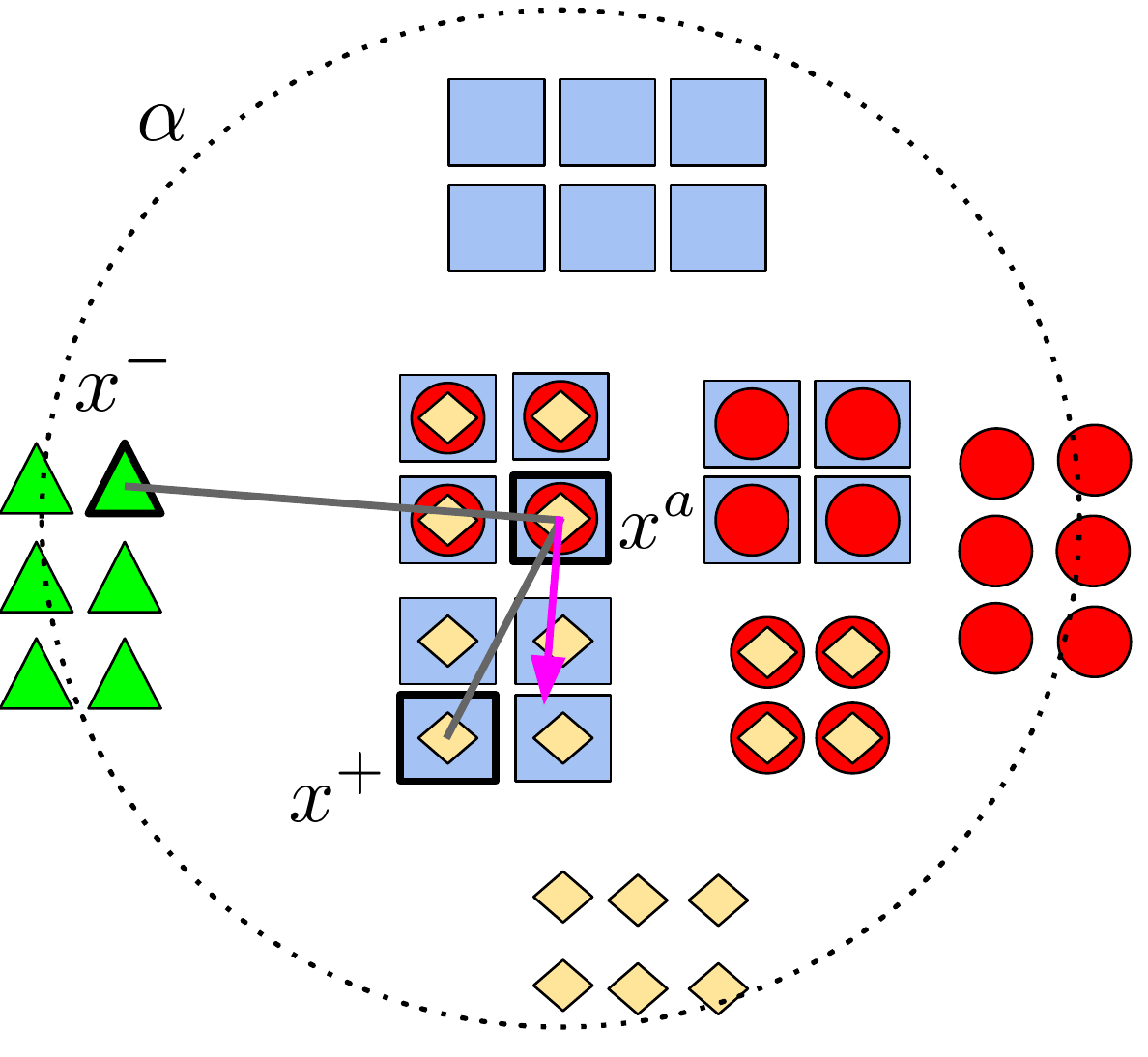}		
			\captionsetup{labelformat=empty}		
			\caption*{a) Triplet loss}
		\end{minipage}%
		\hspace{0.3cm}
		\begin{minipage}[b]{.22\textwidth}
			\centering
			\includegraphics[width=1\linewidth]{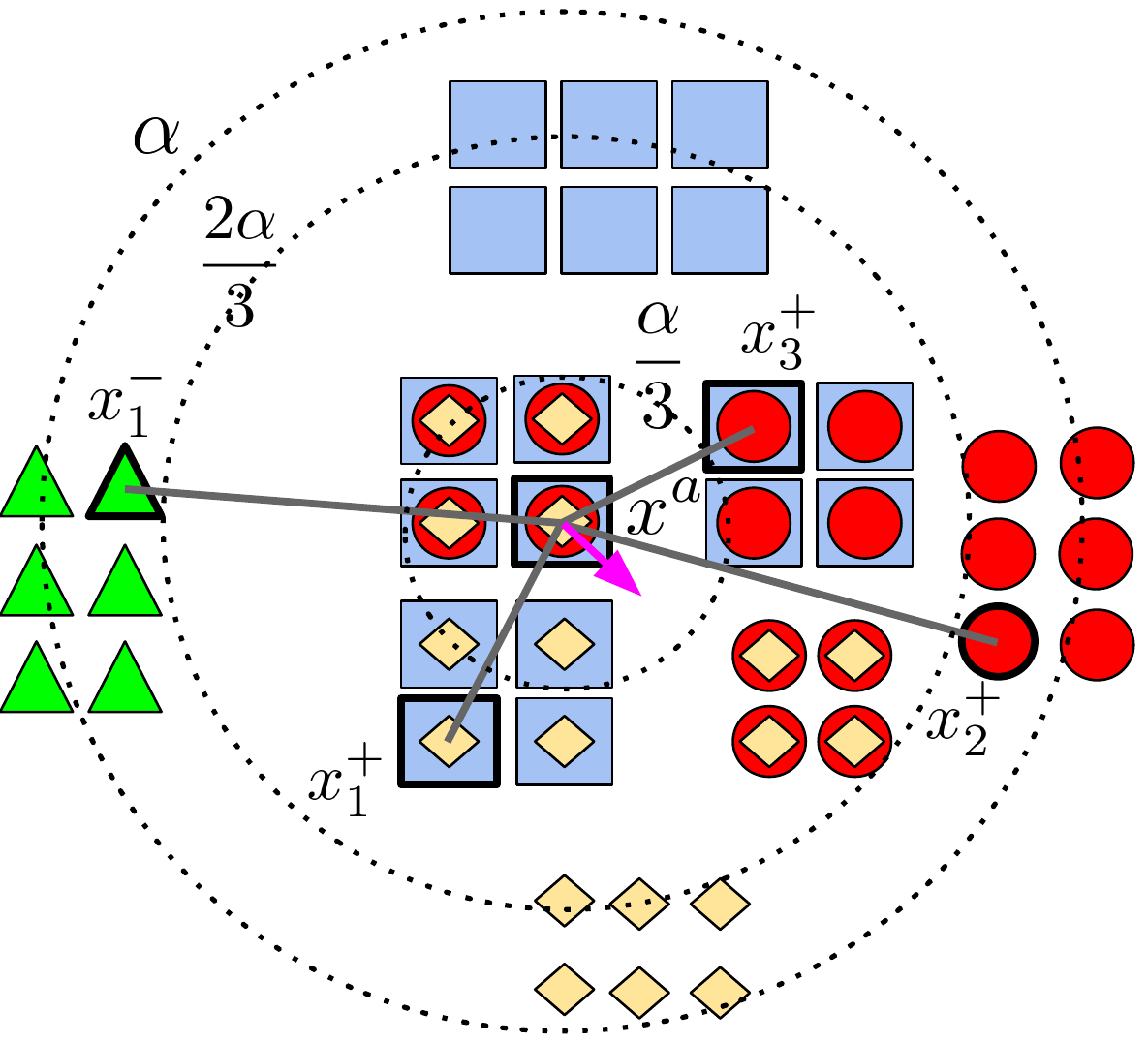}
			\captionsetup{labelformat=empty}
			\caption*{b) ML2 loss}
		\end{minipage}%
		\hspace{0.3cm}
		\begin{minipage}[b]{.22\textwidth}
			\centering
			\includegraphics[width=1\linewidth]{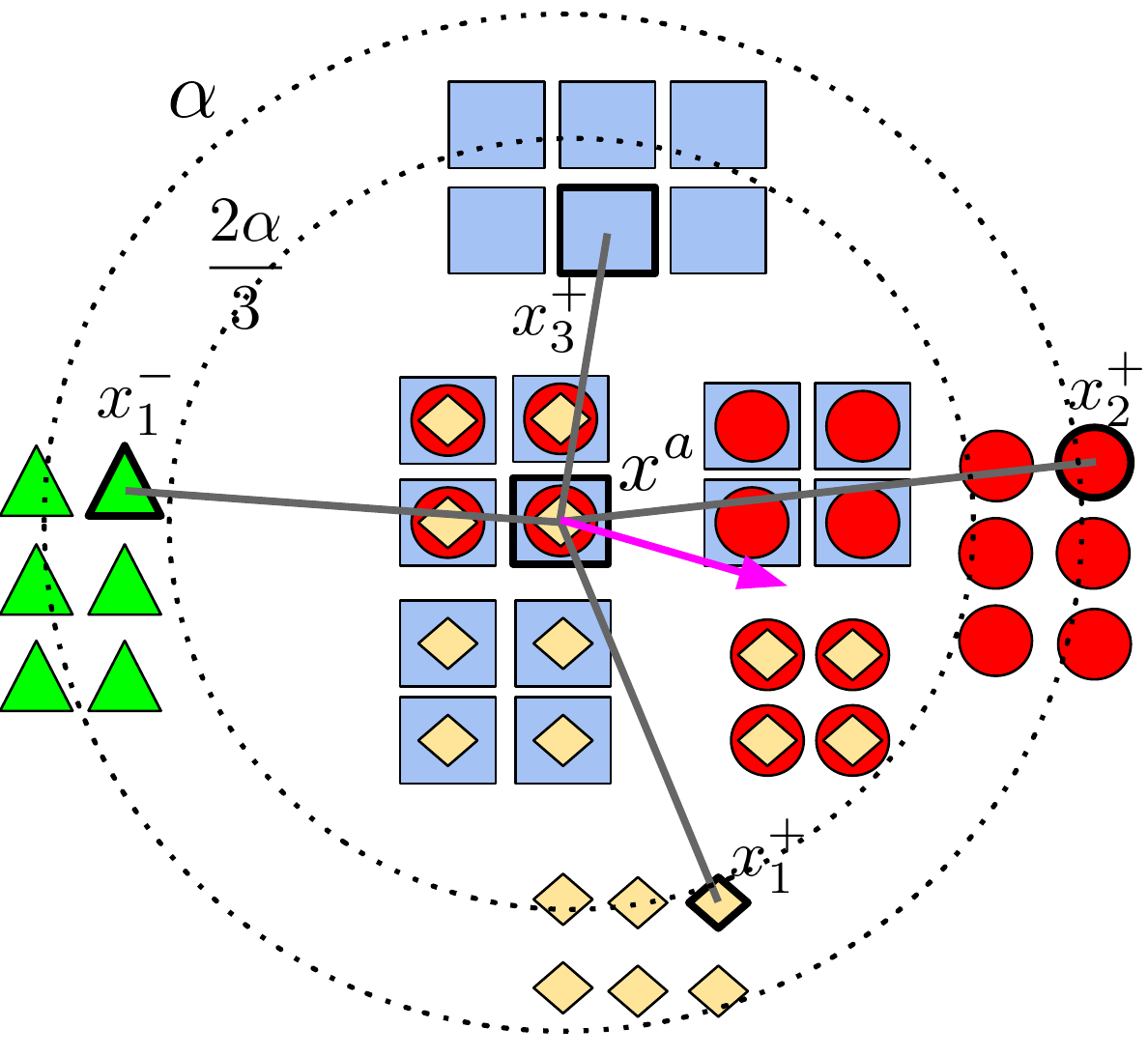}
			\captionsetup{labelformat=empty}
			\caption*{c) ML2+ loss}
		\end{minipage}%
		\hspace{0.3cm}
		\begin{minipage}[b]{.22\textwidth}
			\centering			
			\includegraphics[width=1\linewidth]{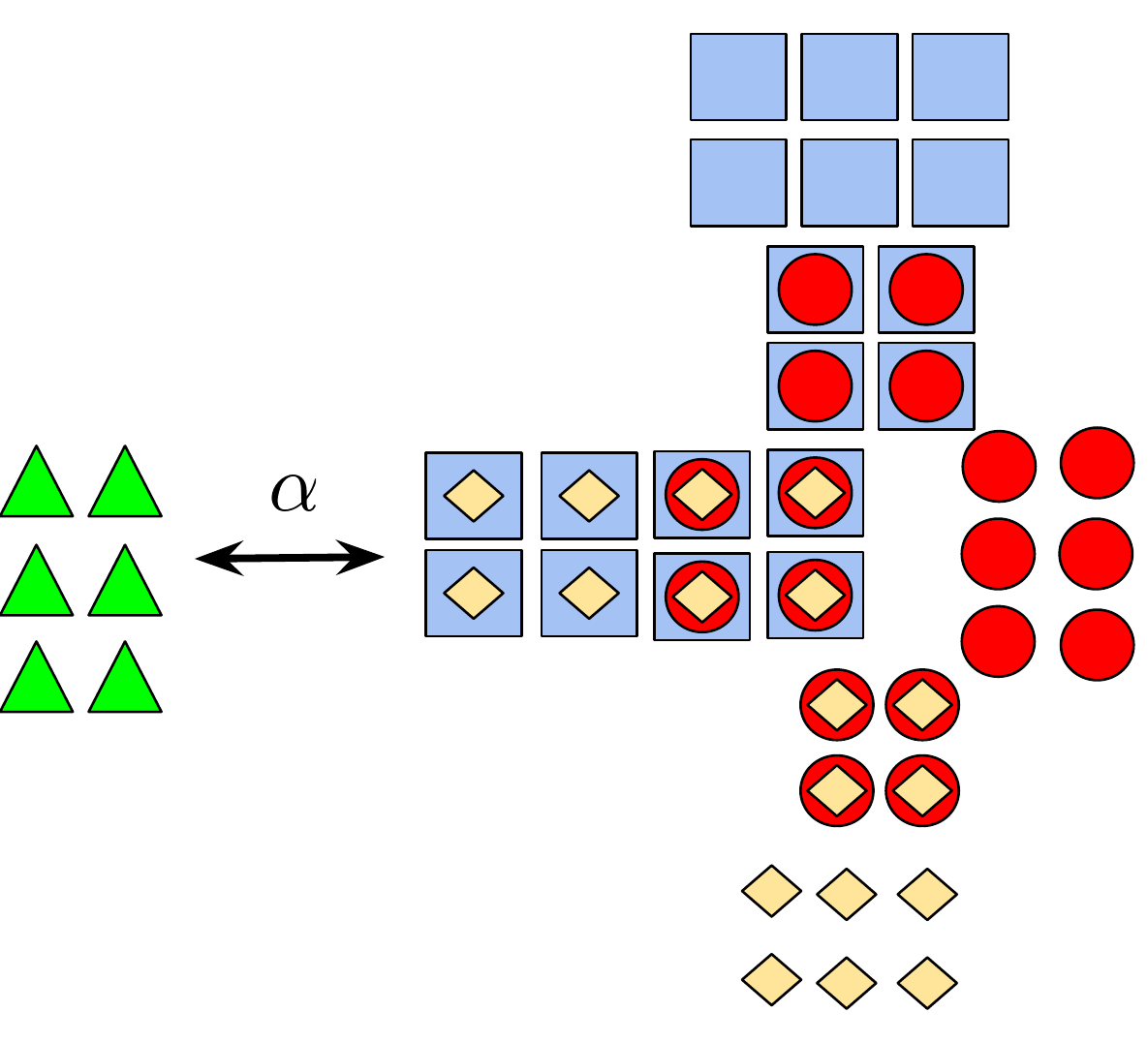}
			\captionsetup{labelformat=empty}
			\caption*{d) Ideal metric}		
		\end{minipage}%
		\caption{
			An illustration of metric learning using the triplet (a), ML2 (b) and ML2+ (c) losses compared to an ideal metric. Each shape represents a label and overlapping shapes indicate co-occurring labels. The dotted arcs indicate the margin bounds depending on $\alpha$. See the text for further details. 
		}
		\label{fig:losses}
	\end{figure*}
	
	\subsection{CAD systems for chest radiographs}
	
	The use of computer-aided diagnosis (CAD) systems in medical imaging goes back more than a half century \cite{van2017fifty}. Over the years the methodologies powering the CAD systems have evolved substantially from rule-based engines to artificial neural nertworks. In recent years, CAD developers have started to adopt deep learning stategies in a number of medical application domains. For instance, Geras et al. \cite{HighRes_BreastCancer_GerasWKMC17} have developed a DCN model able to handle multiple views of high-resolution screening mammographies, which are commonly used to screen for breast cancer. For applications to plain chest radioghraphs, standard DCNs have been used to predict pulmonary tuberculosis \cite{radiology_tuberc_2017} and an architecture involving DCNs and recurrent neural networks has been trained to perform automatic image annotation \cite{ReadingXrays_CVPR2016}. Wang et al. \cite{ChestXray8_WangPLLBS17} have used a database of chest x-rays with more than $100,000$ frontal-view images and  associated radiological reports in an attempt to detect commonly occurring thoracic diseases. 

	\section{Deep metric learning with multi-labelled images}
	
	\subsection{Problem formulation}
	
	In the remainder of this article we assume that each chest radiograph $x \in \mathbb{R}^{w}$  is associated with any of $l$ possible labels contained in a set $\mathcal{L}$. We collect all the labels describing $x$ in a set $\mathcal{L}(x)$ whilst all the remaining labels are identified by $\mathcal{\overline{L}}(x) = \mathcal{L} - \mathcal{L}(x)$. Our aim is to learn a non-linear embedding $f(x)$ that maps each $x$ onto a feature space $\mathbb{R}^d $ where $d \ll w$. In this subspace, the Euclidean distance among groups of similar images should be small and, conversely, the distance between dissimilar images should be large. The distance should be robust to anatomical variability within the normal range as well as geometric distortions and noise. Most importantly, it should be able to capture a notion of {\it radiological similarity}, i.e. two images are expected to be more similar to each other if they share similar radiological abnormalities. We require the embedding function, $f_\theta(\cdot)$, to depend only upon a learnable parameter vector $\theta$. No assumptions about this function can be made besides differentiability with respect to $\theta$. Consequently, the learned distance, $d_\theta(f_\theta(x_i), f_\theta(x_j))$, also depends on $\theta$.
	
	While the definition of \textit{positive} and \textit{negative} elements is straightforward for applications involving mutually exclusive labels, it becomes more ambiguous when each image is allowed to have non-mutually exclusive labels. Restrictive assumptions would need to be made in order to use existing approaches based on contrastive loss \cite{chopra2005learning}, triplet loss \cite{schroff2015facenet} and others \cite{song2015deep,sohn2016improved}. The simplest approach would be to assume that $x_i$ and $x_j$ are \textit{positive} with respect to each other only when they share exactly the same labels, i.e. when $\mathcal{L}(x_i) = \mathcal{L}(x_j)$; conversely, they would be interpreted as negative elements when the equality is not satisfied. However, assuming that two films are radiologically similar when they share exactly the same abnormalities is too strong. 
	Adopting this strategy would also result in much larger sample sizes for elements with frequently co-occuring labels compared to elements characterised by less frequent labels thus hindering the learning process. Furthermore, since each individual label in both $\mathcal{L}(x_i)$ and $\mathcal{L}(x_j)$ is expected to be noisy, requiring the co-occurrence of exactly all the labels may be too restrictive. 
	
	
	A much less restrictive approach would be to assume that $x_i$ and $x_j$ are {\it positive} when they have at least one common label, i.e. when $\mathcal{L}(x_1) \cap \mathcal{L}(x_2) \neq \emptyset$. Under this definition, both the contrastive or triplet loss could still be used. This approach is still far from ideal, though, because this definition is invariant to the degree of overlap between $\mathcal{L}(x_i)$ and $\mathcal{L}(x_j)$. Ideally, the learned distance between any two images should be proportional to the number of abnormalities they do not share. Fig. \ref{fig:losses}d illustrates this ideal situation.  The triplet loss would struggle to satisfy this requirement as it does not take the global structure of the embedding space into consideration \cite{sohn2016improved} and does not explicitly account for overlapping labels; see Fig. \ref{fig:losses}a. In the next section, we propose two loss functions that are designed to overcome the above limitations.
	
	\subsection{Proposed loss functions for multi-labelled images}
	
	We begin by assuming that $x_i$ and $x_j$ are positive when $\mathcal{L}(x_i) \cap \mathcal{L}(x_j) \neq \emptyset$. Given an anchor $x^a$, our approach starts by retrieving $l$ randomly selected images, one for each label in $\mathcal{L}$. The images are then grouped into two non-overlapping sets: one containing $p$ {\it positive} elements $$\mathcal{P}(x^a) = \{x_1^+,...,x_p^+\}$$ and one containing the $n$ remaining negative elements $$\mathcal{N}(x^a) = \{x_1^-,...,x_n^-\}$$ where $p+n=l$.  An ideal metric should ensure that $x^a$ is kept as close as possible to all the elements in $\mathcal{P}$ whilst being kept away from all the elements in $\mathcal{N}$. Accordingly, the loss function to be minimised can be defined as 
	$$ L(x^a,\mathcal{P},\mathcal{N}) = \frac{1}{n p} \sum\limits_{i=1}^{p} \sum\limits_{j=1}^{n} \max \bigg(0,L_{tpl}(x^a,x^+_i,x^-_j) \bigg) $$
	$$ L_{tpl}(x^a,x^+,x^-) = d\big(f_\theta(x^a),f_\theta(x^+)\big) - d\big(f_\theta(x^a),f_\theta(x^-)\big) + \alpha $$
	where the positive scalar $\alpha$ represents a margin to be enforced between positive and negative pairs. This formulation can be seen as the triplet loss average derived from all the possible triplets $\{x^{a},x_i^{+}, x_j^{-} \}$ where $x_i^+ \in \mathcal{P}$ and $x_j^- \in \mathcal{N}$.
	
	The expression above can be simplified by pre-selecting the negative element $x_j^-$ having the largest contribution (e.g. see also Song et al.\cite{song2015deep}), i.e. yielding 
	$$
	L^-(x_a, \mathcal{N}) =  \max_{j} \Big[\alpha - d\big(f_\theta(x^a),f_\theta(x^-_j)\big) \Big]
	$$ 
	In this way, we obtain a more tractable optimisation problem 
	$$ L(x^a,\mathcal{P},\mathcal{N}) = \frac{1}{n p} \sum\limits_{i=1}^{p} \max \bigg(0,d\big(f_\theta(x^a),f_\theta(x^+_i)\big) +  L^-(x_a, \mathcal{N}) \bigg) $$
	which can be further simplified by using a smooth upper bound for $L^-(x_a, \mathcal{N})$,
	$$
	\hat{L}^-(x_a, \mathcal{N}) = \log\Big(  \sum\limits_{j=1}^{n} e^{\alpha - d\big(f_\theta(x^a),f_\theta(x_j^-)\big)}\Big) \geq L^-(x_a, \mathcal{N}) 
	$$
	

	The above loss does not directly address the issue arising when some elements in $\mathcal{P}(x^a)$ have labels that are not in $\mathcal{L}(x^a)$. Without imposing further constraints on how the elements in $\mathcal{P}$ are selected, the loss will force $d\big(f_\theta(x^a),f_\theta(x^+)\big)$ to become as small as possible regardless of the number of labels that $x^a$ and $x^+$ actually have in common. This problem is addressed by introducing a quantity, $\tau$, that represents the degree of overlap between the labels associated to $x^a$ and those associated to its positive elements, i.e. 
	$$
	\tau = \frac{ \big(| \mathcal{L}(x^a) \cup \mathcal{L}(x^+_i) | - | \mathcal{L}(x^a) \cap \mathcal{L}(x^+_i) | \big)}{| \mathcal{L}(x^a) \cup \mathcal{L}(x^+_i)|}.
	$$
	Clearly, $\tau$ is equal to $0$ when $| \mathcal{L}(x^a) \cap \mathcal{L}(x^+_i) | = | \mathcal{L}(x^a) \cup \mathcal{L}(x^+_i) | $ and to $1$ when $\mathcal{L}(x^a) \cap \mathcal{L}(x^+_i) = \emptyset$. By allowing $d\big(f(x^a),f(x_i^+)\big)$ to be a fraction $\tau$ of $\alpha$, we obtain the proposed ML2 (Metric Learning for Multi-Label) loss, i.e.
%
	$$ ML2_{loss} = \frac{1}{p} \sum\limits_{i=1}^{p} \max \bigg(0,  d\big(f_\theta(x^a),f_\theta(x_i^+)\big) - \alpha \tau + \hat{L}^-(x_a, \mathcal{N})  \bigg) $$ 
	An illustrative example of its inner working is provided in Fig. \ref{fig:losses}b. We also propose a different version of the loss, which relies on a different definition of positive elements. In this case, for each label in $\mathcal{L}(x^a)$, a \textit{positive} element is strictly required to have only that particular label. The quantify $\tau$ then simplifies to $\tau =  (p-1)/p $ since $|\mathcal{L}(x^a) \cap \mathcal{L}(x^+_i)| = 1$ and $|\mathcal{L}(x^a) \cup \mathcal{L}(x^+_i)| = p$. An illustration is provided in Fig. \ref{fig:losses}c, and we call this version ML2+. 
	
	
	For applications involving a large number of classes, a memory efficient implementation of the two methods above can be obtained by reducing the elements in $\mathcal{P}$ and $\mathcal{N}$ using a hard class mining approach. In this case, $\mathcal{P}$ and $\mathcal{N}$ depend only on a subset of all $l$ labels, which is chosen by determining which labels contribute the most to the overall loss (e.g. see Sohn et al.\cite{sohn2016improved}).

	\section{Large-scale metric learning for chest radiographs}
	
	\begin{table}[t]
		\begin{center}
			\caption{Dataset sample sizes}
			\label{tab:dataset}
			\begin{tabular}{|c|c|c|c|c|}
				\hline  Class            & Train 	& Validation	& Test		& GL Set \\
				\hline  Normal           & 86863	& 10857			& 10865		& 558    \\
				\hline  Cardiomegaly     & 40312	& 5084			& 5315		& 374    \\
				\hline  Medical device	 & 105880	& 13287			& 13616		& 850    \\
				\hline  Pleural effusion & 66980 	& 8398			& 8676		& 642    \\
				\hline  Pneumothorax	 & 20003 	& 2519			& 2613		& 212    \\
				\hline 
				\hline  Total			 & 261678 	& 32802			& 33494		& 2051    \\
				\hline
			\end{tabular} 
		\end{center}
	\end{table}
	
	For this study, we obtained a large dataset consisting of $745,480$ historical chest radiographs extracted from the PACS system of Guy's \& St Thomas' NHS Foundation Trust, serving a large, diverse population in South London. Our dataset covers the period between January 2005 and March 2016. The radiographs were taken using $40$ different scanners across more than $100$ departments. For a large portion of these exams, we had both the radiological report as well as the associated plain film. The reports were written by $276$ different readers, including consultant and trainee radiologists and accredited reporting radiographers. All the examinations were anonymised with no patient-identifiable data or referral information. The size of the images ranges from $734 \times 734$ to $4400 \times 4400$ pixels, and each pixel is represented in greyscale with 12 bit precision. Table \ref{tab:dataset} contains the sample size breakdown of all the exams that we used for training, validation, and testing. Starting from the full dataset, we selected all the exams concerning patients older than $16$ years and for which we had both the report and the plain film. Only a subset of manually validated exams - the {\it Golden Set} - was used to assess and compare the performance of the metric algorithms.
	
    \subsection{Automatic labels extraction from medical reports}
    
    Given the large number of reports available for the study, obtaining manual labels for each exam was unfeasible. Instead, all the written reports were processed using a NLP system specifically developed to model radiological language \cite{cornegruta2016modelling}. The system was trained to detect any mention of radiological abnormalities and their negations. Labels were chosen to allow all common radiological findings to be allocated to a group along with other films sharing similar appearances. The labels were adapted from Hansell et al. \cite{hansell2008fleischner} and were meant to capture discrete radiological findings (e.g. cardiomegaly, medical device, pleural effusion) rather than giving a final diagnosis (e.g. pulmonary oedema), which requires clinical judgement to combine the current findings with previous imaging, clinical history, and laboratory results. For this study, we used $l=4$ different labels, i.e. {\it cardiomegaly}, {\it medical devices} (e.g. pacemakers, lines, and tubes), {\it pleural effusion} and {\it pneumothorax}. The NLP system also identified all ``normal'' exams, i.e. those where no abnormalities were mentioned in the report. Cumulatively, the normal and abnormal labels used here represent $68\%$ of all the reported visual patterns in our database. 
	
	A validation study was carried out to assess how accurately the NLP system extracted the $4$ clinical labels, plus the normal class, from the written reports. Two independent clinicians were presented with the original radiological reports and manually generated the labels from the reports. This study generated the \textit{Golden Set}, which is used here purely for performance evaluation purposes. 
	
	In Table \ref{tab:NLP_performace} we report the precision, sensitivity, specificity and $F1$ score obtained by our NLP system. These results demonstrate that the labels automatically extracted at scale from the written reports are sufficiently reliable; this provides evidence that the vast majority of labels associated to images in our datasets is correct, thus allowing the neural network architectures to learn suitable image representations. 
	
	Further details on the NLP algorithms and experimental results can be found in Pesce et al. \cite{2017arXiv171200996P}.

	\begin{table}
		\begin{center}
			\caption{NLP performance on the \textit{Golden Set}}
			\label{tab:NLP_performace}
			\begin{tabular}{|c|c|c|c|c|}
				\hline  Class            & Prec.	& Sens.		& Spec.		& $F_1$		\\
				\hline  Normal           & 98.98	& 97.33		& 99.85		& 98.15		\\
				\hline  Cardiomegaly     & 99.59	& 99.39		& 99.95		& 99.49		\\
				\hline  Medical device	 & 98.52	& 94.34		& 99.27		& 96.39		\\
				\hline  Pleural effusion & 96.80	& 91.42		& 99.36		& 94.03		\\
				\hline  Pneumothorax	 & 77.07	& 96.88		& 98.05		& 85.85		\\
				\hline  
			\end{tabular} 
		\end{center}
	\end{table}

	\subsection{DCN for high resolution input images}
	
	Standard DCN architectures, such as \textit{Inception v3}, were originally designed to model natural images, such as those in the Imagenet dataset \cite{ILSVRC15}. These images are typically scaled down to $299\times299$ pixels, even though higher resolution images are available. In many studies, down-scaling natural images has been shown to be a good compromise between the amount of information that is lost and computational efficiency. However, in a medical imaging setting, every detail in an image matters, at least in principle. Thus, arbitrarily reducing the resolution of the images is generally considered suboptimal \cite{HighRes_BreastCancer_GerasWKMC17}. For this reason, in our study we have implemented a slightly modified version of \textit{Inception v3} that is able to handle $1211\times1083$ pixels images. Table \ref{tab:inceptionv3_1211x1083_arc} shows the details of the proposed architecture. The chosen aspect ratio is close to the median aspect ratio ($13:12$) amongst all images in our dataset and has the advantage of minimizing the number of image that would be cropped (or padded), since the input of our DCN has a fixed size.
		
	\begin{table}
		\begin{center}
			\caption{Proposed architecture based on the \textit{Inception v3} network for $1211\times1083$ pixels input images.}
			\label{tab:inceptionv3_1211x1083_arc}
			\begin{tabular}{lccc}
				\hline  
				\multicolumn{1}{|l|}{\Centerstack[l]{module}}	& \multicolumn{1}{c|}{\Centerstack[c]{patch size / stride \\ (padding)}}	& \multicolumn{1}{c|}{\Centerstack[c]{rep.}}	& \multicolumn{1}{c|}{\Centerstack[c]{output size}}	\\ \hline
				\hline
				\multicolumn{1}{|l|}{conv}		& \multicolumn{1}{c|}{$3\times3$ / 2}		& \multicolumn{1}{c|}{$\times 1$}				& \multicolumn{1}{c|}{$32\times605\times541$}	\\ \hline
				\multicolumn{1}{|l|}{conv}		& \multicolumn{1}{c|}{$3\times3$ / 1}		& \multicolumn{1}{c|}{\multirow{3}{*}{$\times 4$}} & \multicolumn{1}{c|}{\multirow{3}{*}{$48\times301\times269$}}	\\ \cline{1-2}
				\multicolumn{1}{|l|}{conv}		& \multicolumn{1}{c|}{$3\times3$ / 1 ($+1$)}	& \multicolumn{1}{c|}{}& \multicolumn{1}{c|}{}	\\ \cline{1-2}
				\multicolumn{1}{|l|}{max pool}	& \multicolumn{1}{c|}{$3\times3$ / 2}		& \multicolumn{1}{c|}{}& \multicolumn{1}{c|}{}	\\ \hline			
				\multicolumn{1}{|l|}{Incep. A}		& \multicolumn{1}{c|}{\multirow{3}{*}{\shortstack{See Szegedy \\ et al. \cite{szegedy2015rethinking}}}}		& \multicolumn{1}{c|}{$\times 3$}				& \multicolumn{1}{c|}{$1088\times17\times15$}	\\ \cline{1-1} \cline{3-4}	
				\multicolumn{1}{|l|}{Incep. B}		& \multicolumn{1}{c|}{}		& \multicolumn{1}{c|}{$\times 5$}				& \multicolumn{1}{c|}{$1600\times8\times7$}	\\ \cline{1-1} \cline{3-4}
				\multicolumn{1}{|l|}{Incep. C}		& \multicolumn{1}{c|}{}		& \multicolumn{1}{c|}{$\times 2$}				& \multicolumn{1}{c|}{$2048\times8\times7$}	\\ \hline
				\multicolumn{1}{|l|}{avg. pool}		& \multicolumn{1}{c|}{$8\times7$ / 1}		& \multicolumn{1}{c|}{$\times 1$}				& \multicolumn{1}{c|}{$2048\times1\times1$}	\\ \hline												
			\end{tabular} 
		\end{center}
	\end{table}
	
	\section{Experimental results}

	\subsection{Training strategy}
	
	The $f_\theta(x)$ representation was learned using an \textit{Inception v3} architecture \cite{szegedy2015rethinking} resulting in an $m-$dimensional mapping under the constraint that $\lVert f_\theta(x)\rVert_2 = 1 $. We call $g_\psi(x)$ the output of the last convolutional layer and we define our final layer as:
	$$ f_\theta(x) = \dfrac{g_\psi(x)\beta + b}{\norm{g_\psi(x)\beta + b}^2 } $$
	where $\beta \in \mathbb{R}^{2048 \times m}$ and $b \in \mathbb{R}^{2048}$ are respectively weights and bias of the last layer.
	All the results presented here use $m=64$, because the use of larger dimensions did not introduce any significant improvements.
	All images were rescaled to have a standard size of $299 \times 299$ ($1211 \times 1083$ for the non-standard model) pixels and no other pre-processing was carried out. For training purposes, synthetic data was generated by random rotation and flipping of the original images. 
	Two different experiments setups were considered, one in which the $f_\theta(x)$ was learned end-to-end from the raw images, and one where pre-training was used instead, as is commonly done in other works\cite{sohn2016improved}\cite{song2015deep}. The proposed ML2 and ML2+ losses were compared to more traditional metric learning approaches based on contrastive and triplet losses sharing the same architecture.
	
	Stochastic Gradient Descent (SGD) was used for the optimisation process, with an initial learning rate equal to $0.01$, momentum equal to $0.9$ and weight decay equal to $10^{-4}$.
	When we started from randomly initialised weights, the total number of iterations was $90,000$ and, every $25,000$ iterations, the learning rate was decreased by a factor of $10$.
	Instead, when the weights were pre-trained on the classification task, the number of total iterations was $27,000$ and the learning rate was decreased every $8,000$ iterations.
	In both experimental setups the size of the mini-batches was equal to $36$ when contrastive and triplet losses were used, and it was equal to $10$ for our proposed losses. We tested different values for $\alpha$, which, for the results shown in this work, has been set to $0.2$. During the training the model with the best value of NMI on the validation set is kept as the best model and used during the testing phase.	

	Positive and negative elements were randomly sampled. The noisiness of our labels prevented us from exploiting any sampling techniques (e.g. hardest negative mining, etc.), since all those methods take the reliability of the labels for granted. 	
	
	\subsubsection{DCN pre-training}

	For the pre-training of our DCN, we used a multi-label binary cross entropy loss. Given our $4$ possible labels, we defined an equal number of binary classifiers with the aim of predicting the presence or absence of each label. 
	The output of each binary classifier $ l^i_\phi(x)$ is
	$$ l^i_\phi(x) =  LogSoftMax\big( g_\psi(x)\beta_i + b_i \big) $$
	where $\beta_i \in \mathbb{R}^{2048 \times 2}$ and $b_i \in \mathbb{R}^{2048}$ are different weights and bias whith respect to the one defined above.
	The loss function is equal to the average of the negative log likelihoods of $l^i_\phi(x)$ for each $i=1,... ,4$,
	$$L(l_\phi(x),y) =  \frac{1}{4} \sum\limits_{i=1}^{4} y_{i} \cdot l^{i}_\phi(x) $$
	where $y$ is the labels vector; $y_i$ will be equal to $(1,0)$ when the i-th abnormality is present in the image $x$, otherwise, it will be equal  to $(0,1)$ .
	
	\subsection{Cluster and retrieval performance}

	We assessed the performance of the proposed losses on two different tasks: (i) clustering, evaluated with the normalized mutual information (NMI) metric and (ii) image retrieval, evaluated with the Recall@K metric; see Manning et al. \cite{manning2008introduction} for a complete account of these metrics.

	Table \ref{tab:results_on_gls} shows the empirical results obtained after learning the metrics on the $ 263,513 $ training images and testing them on the Golden Set. When learning without pre-training (i.e. initially using random weights), ML2+ outperforms ML2 on both tasks and largely improves upon the other alternative losses. When using a pre-trained architecture, improvements can be observed across all methods, and ML2+ obtains a slightly better performance than ML2. Based on these results, we demonstrate the superior performance of our proposed losses with respect to the baseline; moreover, we suspect that ML2+ is able to converge to a better optimum more easily than ML2.
	
	In the same Table we also reported the results obtained with a DCN using $1211\times1083$ pixels input images. Here we used the same configuration of the model yielding the best performance on standard image size (i.e. pre-trained weights and ML2+ loss). Almost no improvements at all can be seen compared to the standard version of the network. In fact, while the retrieval performances are almost the same, the NMI score is more than one point lower. We hypothesize that, at least for the radiological abnormalities we have considered here, which involve large anatomical structures, an input size of $299\times299$ pixels may be sufficiently informative.
	
\begin{sidewaysfigure*}[t!]
		\centering
		\includegraphics[scale=1.0]{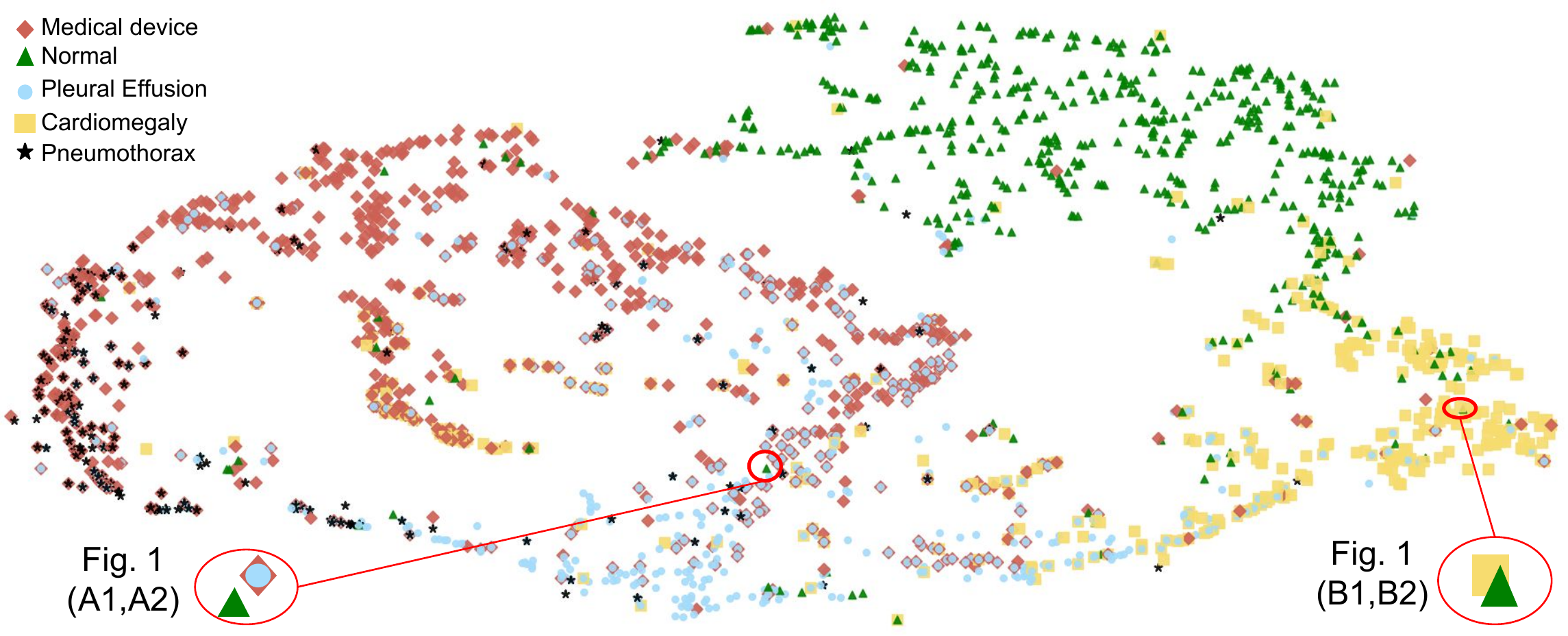}
		\caption{$2$-dimensional embedding of all chest radiographs contained in the golden dataset learned through the ML2+ loss and visualised via multi-dimensional scaling. Each exam is represented as a point with different shapes and colors to identify multiple labels. Well-separated cluster of ``normal'' radiographs (green triangles) and exams featuring an enlarged heart are clearly visible. See Fig. \ref{fig:x-rays} for the circled images.}
		\label{fig:TSNE}
	\end{sidewaysfigure*}

	Figure \ref{fig:TSNE} shows a $2$-dimensional representation of the $2,252$ radiographs contained in the \textit{Golden Set}. This representation was obtained by means of dimensionality reduction using a $t$-distributed Stochastic Neighbor Embedding (t-SNE) \cite{JMLR_vandermaaten14a}, which effectively projects the $64$-dimensional embeddings extracted from the best model onto $2$ dimensions for visualisation purposes. Remarkably, this projection shows that the normal exams are mostly concentrated in a well-separated cluster; moreover, other clusters of exams sharing similar abnormalities have also been identified. 
	
	The chest radiographs marked with a circle can be seen in Figure \ref{fig:x-rays}. These are two examples of radiographs that were originally labelled as normal but ended up being placed away from the cloud of normal exams. A second reading of these exams has revealed unreported abnormalities thus confirming that their position within the embedding was justified. 
	
	\begin{table*}
		\begin{center}	
			\begin{tabular}{cc|c|c|c|c|c|c|c|c|c|c|c|}
				\cline{3-7} \cline{9-13}
				&& \multicolumn{5}{c|}{Without pre-training} &  & \multicolumn{5}{c|}{With pre-training} \\  \cline{3-7} \cline{9-13} 
					
				&& NMI				& R@1				& R@2				& R@4				& R@8 
				&& NMI				& R@1				& R@2				& R@4				& R@8            
				\\ \cline{1-1} \cline{3-7} \cline{9-13}
				  
				\multicolumn{1}{|c|}{Contrastive}    
				&& 17.57			& 32.86				& 46.76				& 60.90				& 74.65
				&& 37.76			& 52.71				& 64.16				& 74.99				& 84.11          
				\\ \cline{1-1} \cline{3-7} \cline{9-13}	
						                              
				\multicolumn{1}{|c|}{Triplet}        
				&& 27.24			& 41.30				& 55.58				& 69.58				& 81.08
				&& 39.46			& 52.22				& 66.02				& 78.25				& 86.49           
				\\ \cline{1-1} \cline{3-7} \cline{9-13}	
					                              
				\multicolumn{1}{|c|}{ML2 }     
				&& 35.79			& 47.05				& 62.21				& 76.26				& 84.84
				&& \textbf{41.76}	& 53.83				& 67.19				& 78.64				& 86.40           
				\\ \cline{1-1} \cline{3-7} \cline{9-13}
				
				\multicolumn{1}{|c|}{ML2+ }    
				&& \textbf{37.32}	& \textbf{50.51}	& \textbf{64.90}	& \textbf{78.25}	& \textbf{86.74}
				&& 41.62			& \textbf{54.27}	& \textbf{67.28}	& \textbf{79.18}	& \textbf{87.37}           
				\\ \cline{1-1} \cline{3-7} \cline{9-13}							
				
				\\ \cline{1-1} \cline{3-7} \cline{9-13}
				\multicolumn{1}{|c|}{ML2+ (high res.)}    
				&& --				& --				& --				& --				& --
				&& 40.80			& 54.90				& 68.11				& 79.62				& 86.64        
				\\ \cline{1-1} \cline{3-7} \cline{9-13}				
			\end{tabular}

		\end{center}
		\caption{Clustering and retrieval results of different metric learning losses in terms of NMI and R@1,2,4,8.}
		\label{tab:results_on_gls}
	\end{table*}
	
	\subsection{Abnormalities classification performance}
	
	\begin{table}
		\begin{center}
			\caption{The classification performance for abnormal exams obtained (i) when the network is trained directly on the classification task (Cross-entropy),   (ii) using the embeddings extracted from a network trained with a triplet loss in order to train a linear regression classifier (LR on triplet embedding) and (iii) using the embeddings extracted from a network trained with our proposed loss, ML2+, in order to train a linear regression classifier (LR on ML2+ embedding).}
			\label{tab:CV_abnormal_class}
			\begin{tabular}{|c|c|c|c|c|}
				\hline  Method						& Prec.				& Sens.				& Spec.				& $F_1$				\\
				\hline  Cross-entropy				& 94.86				& \textbf{95.24}	& 86.20				& 95.05				\\
				\hline  LR on triplet embedding		& 95.36				& 95.04				& 87.63				& 95.20				\\
				\hline  LR on ML2+ embedding		& \textbf{95.55}	& 94.98				& \textbf{88.17}	& \textbf{95.26}	\\
				\hline  
			\end{tabular} 
		\end{center}
	\end{table}
	
In a separate task, we tried to predict whether a given chest radiograph contains a radiological abnormality. For this task, we compared the performance of the DCN architecture trained as a multi-label classifier using a cross entropy loss (the same described above and used for pre-training) and the feature embeddings extracted from one of our DCN trained with a metric loss. Logistic regression was used on the extracted embedding space in order to obtain a classification prediction. In Table \ref{tab:CV_abnormal_class} we present the results we obtained. Performances are evaluated in terms of Precision, Sensitivity, Specificity and $F_1$ Score. We used $F_1$ Score instead of Accuracy bacause in our data normal and abnormal exams are not balanced, and in the latter case comparing performances using Accuracy can be misleading. In comparison to the baseline model, it is possible to see that the models based on the learned embedding obtain better performances, showing a higher proficiency when discriminating between normal and abnormal exams.

	\section{Conclusions}
	
	In this article we have proposed two loss functions for metric learning with multi-labelled medical images. Their performance has been tested on a very large dataset of chest radiographs. Our initial results demonstrate that learning a metric that captures a notion of radiological similarity is indeed possible; most importantly, the learned metric places normal radiographs far away from the exams that have been reported to contain one or multiple abnormalities. This is a striking result, given the complexity of the visual patterns to be discovered, the degree of noise characterising the radiological labels, and the large variety of scanners and readers included in our study. It is also an important step towards the fully-automated reading of chest radiographs as being able to recognize normal radiological structures on plain film, which is key to interpreting any abnormal findings.

\section*{Acknowledgments}
The authors thank NVIDIA for providing access to a DGX-1 server, which speeded up the training and evaluation of all the deep learning algorithms used in this work.

%
\bibliographystyle{abbrv}
\bibliography{sigproc}  

\begin{thebibliography}{10}

\bibitem{NIPS1993_769}
J.~Bromley, I.~Guyon, Y.~LeCun, E.~S\"{a}ckinger, and R.~Shah.
\newblock Signature verification using a ``siamese" time delay neural network.
\newblock In {\em NIPS}, pages 737--744. 1994.

\bibitem{chopra2005learning}
S.~Chopra, R.~Hadsell, and Y.~LeCun.
\newblock Learning a similarity metric discriminatively, with application to
  face verification.
\newblock In {\em CVPR}, volume~1, pages 539--546. IEEE, 2005.

\bibitem{cornegruta2016modelling}
S.~Cornegruta, R.~Bakewell, S.~Withey, and G.~Montana.
\newblock Modelling radiological language with bidirectional long short-term
  memory networks.
\newblock {\em 7th International Workshop on Health Text Mining and Information
  Analysis}, 2016.

\bibitem{HighRes_BreastCancer_GerasWKMC17}
K.~J. Geras, S.~Wolfson, S.~G. Kim, L.~Moy, and K.~Cho.
\newblock High-resolution breast cancer screening with multi-view deep
  convolutional neural networks.
\newblock {\em CoRR}, abs/1703.07047, 2017.

\bibitem{hansell2008fleischner}
D.~M. Hansell, A.~A. Bankier, H.~MacMahon, T.~C. McLoud, N.~L. Muller, and
  J.~Remy.
\newblock Fleischner society: glossary of terms for thoracic imaging 1.
\newblock {\em Radiology}, 246(3):697--722, 2008.

\bibitem{radiology_tuberc_2017}
P.~Lakhani and B.~Sundaram.
\newblock Deep learning at chest radiography: Automated classification of
  pulmonary tuberculosis by using convolutional neural networks.
\newblock {\em Radiology}, 284(2):574--582, 2017.
\newblock PMID: 28436741.

\bibitem{manning2008introduction}
C.~D. Manning, P.~Raghavan, and H.~Sch{\"u}tze.
\newblock {\em Introduction to Information Retrieval}.
\newblock Cambridge University Press, Cambridge, UK, 2008.

\bibitem{2017arXiv171200996P}
E.~{Pesce}, P.-P. {Ypsilantis}, S.~{Withey}, R.~{Bakewell}, V.~{Goh}, and
  G.~{Montana}.
\newblock {Learning to detect chest radiographs containing lung nodules using
  visual attention networks}.
\newblock {\em ArXiv e-prints}, Dec. 2017.

\bibitem{ILSVRC15}
O.~Russakovsky, J.~Deng, H.~Su, J.~Krause, S.~Satheesh, S.~Ma, Z.~Huang,
  A.~Karpathy, A.~Khosla, M.~Bernstein, A.~C. Berg, and L.~Fei-Fei.
\newblock {ImageNet Large Scale Visual Recognition Challenge}.
\newblock {\em International Journal of Computer Vision (IJCV)},
  115(3):211--252, 2015.

\bibitem{schroff2015facenet}
F.~Schroff, D.~Kalenichenko, and J.~Philbin.
\newblock Facenet: A unified embedding for face recognition and clustering.
\newblock In {\em CVPR}, pages 815--823, 2015.

\bibitem{ReadingXrays_CVPR2016}
H.~Shin, K.~Roberts, L.~Lu, D.~Demner{-}Fushman, J.~Yao, and R.~M. Summers.
\newblock Learning to read chest x-rays: Recurrent neural cascade model for
  automated image annotation.
\newblock In {\em CVPR}, pages 2497--2506, 2016.

\bibitem{sohn2016improved}
K.~Sohn.
\newblock Improved deep metric learning with multi-class n-pair loss objective.
\newblock In {\em NIPS}, pages 1849--1857, 2016.

\bibitem{Song_CVPR2017}
H.~O. Song, S.~Jegelka, V.~Rathod, and K.~Murphy.
\newblock Deep metric learning via facility location.
\newblock In {\em CVPR}, 2017.

\bibitem{song2015deep}
H.~O. Song, Y.~Xiang, S.~Jegelka, and S.~Savarese.
\newblock Deep metric learning via lifted structured feature embedding.
\newblock In {\em CVPR}, 2016.

\bibitem{szegedy2015rethinking}
C.~Szegedy, V.~Vanhoucke, S.~Ioffe, J.~Shlens, and Z.~Wojna.
\newblock Rethinking the inception architecture for computer vision.
\newblock In {\em CVPR}, pages 2818--2826.

\bibitem{JMLR_vandermaaten14a}
L.~van~der Maaten.
\newblock Accelerating t-sne using tree-based algorithms.
\newblock {\em Journal of Machine Learning Research}, 15:3221--3245, 2014.

\bibitem{van2017fifty}
B.~van Ginneken.
\newblock Fifty years of computer analysis in chest imaging: rule-based,
  machine learning, deep learning.
\newblock {\em Radiological Physics and Technology}, 10(1):23--32, 2017.

\bibitem{ChestXray8_WangPLLBS17}
X.~Wang, Y.~Peng, L.~Lu, Z.~Lu, M.~Bagheri, and R.~M. Summers.
\newblock Chestx-ray8: Hospital-scale chest x-ray database and benchmarks on
  weakly-supervised classification and localization of common thorax diseases.
\newblock {\em CoRR}, abs/1705.02315, 2017.

\bibitem{NIPS2005_2795}
K.~Q. Weinberger, J.~Blitzer, and L.~K. Saul.
\newblock Distance metric learning for large margin nearest neighbor
  classification.
\newblock In Y.~Weiss, P.~B. Sch\"{o}lkopf, and J.~C. Platt, editors, {\em
  Advances in Neural Information Processing Systems 18}, pages 1473--1480. MIT
  Press, 2006.

\bibitem{samplingmatters_WuMSK17}
C.~Wu, R.~Manmatha, A.~J. Smola, and P.~Kr{\"{a}}henb{\"{u}}hl.
\newblock Sampling matters in deep embedding learning.
\newblock {\em CoRR}, abs/1706.07567, 2017.

\end{thebibliography}
\end{document}